\documentclass{article}

\usepackage{graphicx}
\usepackage{amsmath}

\title{Discovery of missing disease spreader}

\author{Yoshiharu Maeno \\ Social Design Group \\ email: maeno.yoshiharu@socialdesigngroup.com}

\begin{document}

\maketitle

\begin{abstract}
{\em Is it possible to discover a local outbreak of an infectious disease in a region for which data are missing, but which is at work as a disease spreader?} Node discovery for the spread of an infectious disease is defined as discriminating between the nodes which are neighboring to a missing disease spreader node, and the rest, given a dataset on the number of cases. The spread is described by stochastic differential equations. A perturbation theory quantifies the impact of the missing spreader on the moments of the number of cases. Statistical discriminators examine the mid-body or tail-ends of the probability density function, and search for the disturbance from the missing spreader. They are tested with computationally synthesized datasets, and applied to the SARS outbreak and flu pandemic.
\end{abstract}

\section{Introduction}
\label{problem}

No sooner had a new year begun in 2003 than citizens were seized with panic in Guangdong in south China. Hundreds were suffered from a pneumonia-like strange disease, some of which had been dead. Both Chinese government and Chinese media remained silent all the time as to the risk of a possible epidemic. No one in the rest of the world knew there was any real cause for alarm. But in March, local outbreaks of a mysterious disease were reported in Hong Kong and Southeast Asian countries. The World Health Organization (WHO) issued a global alert. Even then, health authorities could not reveal where the disease had come from. This story at the onset of the Severe Acute Respiratory Syndrome (SARS) outbreak poses an interesting question. Is it possible to discover the presence of a missing disease spreader from the surveillance records on the cases in other regions? This study addresses such a node discovery problem for the spread of an infectious disease.

The spread of an infectious disease is a stochastic phenomenon in a spatially heterogeneous medium. Many mathematical models of disease transmission \cite{Ril07} rely on an epidemiological compartment model and a meta-population network model in formulating stochasticity and spatial heterogeneity \cite{Dan09}, \cite{Sim08}. These models are described by a set of stochastic differential equations \cite{Huf04}. The analysis of the spread includes many tasks from reproduction to prediction. Reproduction of the actual spread \cite{Chr10}, \cite{Ish10} is essential in understanding the role of a so-called super-spreader \cite{Fuj07}, \cite{Sma06} and epidemiological thresholds \cite{Par10}, \cite{Col07}. This is a forward analysis. An inverse analysis includes estimating the transmission parameters from observation \cite{Wal10}, \cite{Kee04}. Network profiling estimates the effectively decisive topology of a transportation network which governs the spread \cite{Mae10}. Network inference in a communication network \cite{Rab08} is a similar problem. Statistical learning and computational Monte-Carlo simulation contribute to developing a reliable bio-surveillance system in a noisy environment \cite{Rei03}, detecting abnormal events \cite{Tak06} as an omen of the outbreak \cite{Lu10}, and predicting the spread to aid the health authorities in containing the outbreak \cite{Col06}. None of these, however, addresses the problem. Node discovery problem for a social network \cite{Mae09} is discovering a person who does not appear in the logs on communication, but actually influential to others in an organization. The substantial nature of the problem is the same as this. But the mathematical solution should be different entirely because of the difference in the mechanism of the spread.

In this study, a perturbation theory quantifies how the perturbation which a missing spreader node exerts disturbs the growth of the number of cases. The missing spreader may be the place where the first patient appears (an index node) or not (an intermediate node). Its presence impedes the reproduction of the observed spread by an unperturbed mathematical model. The irreproducibility is rather the clue to solve the problem. Two statistical discriminators are invented. Their role is discriminating between the nodes which are neighboring to the missing spreader node, and the rest (non-neighboring nodes), given a dataset. It is a collection of the time sequence data on the number of cases, or on the cumulative number of new cases, at the nodes within the scope of surveillance in the early growth phase of the spread. The network topology and transmission parameters may be given as an input to the discriminators. Or they may be unknown and must be estimated. The discriminators examine the mid-body or the tail-ends of the probability density function of the number of cases, and search for disturbed time sequence data to which the perturbation gives rise. The mid-body discriminator is founded on the Kolmogorov-Smirnov test. The tail-end discriminator is founded on the Chauvenet rejection test. The discriminators are tested with a number of computationally synthesized datasets, and applied to the WHO datasets on the SARS outbreak and on the flu pandemic (H1N1 swine influenza A) in 2009.

\section{Problem}
\label{problemDefinition}

The mathematical model of the spread in this study is a special case of a stochastic reaction-diffusion process, the integration of a standard epidemiological SIR compartment model and a meta-population network model. The meta-population network model \cite{Bar08} sub-divides the entire population into distinct sub-populations in many geographical regions. The geographical regions are the nodes $n_{i}\ (i=0, 1, \cdots)$. The transportation between two regions is a pair of unidirectional links. The adjacency matrix $\mbox{\boldmath{$l$}}$, whose $i$-th row and $j$-th column element is $l_{ij}$, determines the network topology. If a link from $n_{i}$ to $n_{j}$ is present, $l_{ij}$ is $1$. If absent, $l_{ij}$ is $0$. The parameter $\mbox{\boldmath{$\gamma$}}$ is a matrix whose $i$-th row and $j$-th column element $\gamma_{ij}$ is the probability at which a person moves from $n_{i}$ to $n_{j}$ per a unit time. In many application areas, an empirical law $\gamma_{ij}=\Gamma_{ij}(\mbox{\boldmath{$l$}})$ determines $\mbox{\boldmath{$\gamma$}}$ as a function of $\mbox{\boldmath{$l$}}$. In the following, $\mbox{\boldmath{$\gamma$}}$ and $\mbox{\boldmath{$l$}}$ are interchangeable.

In the SIR compartment model \cite{Kee08} the state of a person changes from a susceptible state, through an infectious state, to a {\em removed (recovered)} state. The quantity $S_{i}(t)$ is the number of susceptible persons at a node $n_{i}$ at time $t$. $I_{i}(t)$ is the number of infectious persons. $R_{i}(t)$ is the number of recovered persons. $J_{i}(t)$ is the cumulative number of new cases until $t$. The parameter $\alpha$ represents the probability at which an infectious person contacts a person and infect the person per a unit time. The parameter $\beta$ represents the probability at which an infectious person recovers per a unit time. These transmission parameters do not depend on time and sub-populations. The reproductive ratio $r$ is defined by $r=\alpha/\beta$ \cite{Lip03}. Movement, infection and recovery are Markovian stochastic processes which are governed by $\mbox{\boldmath{$\theta$}} = \{ \alpha, \beta, \mbox{\boldmath{$l$}} \}$.

Node discovery is a problem to determine whether the nodes $n_{0}, n_{1}, \cdots, n_{N-1}$ which appear in a given dataset is neighboring to a missing spreader node $n_{N}$ or not. The neighboring nodes are those having links to $n_{N}$. The dataset is either $I_{i}(t_{d})$, or $\Delta J_{i}(t_{d}) = J_{i}(t_{d+1}) - J_{i}(t_{d})$, where $i=0,1,\cdots,N-1$ denotes the nodes, and $d=0,1,\cdots,D-1$ denotes observations. The interval between observations is $\Delta t = t_{d+1}-t_{d}$. The number of data is $ND$. The network topology and the transmission parameters $\mbox{\boldmath{$\theta$}}=\{\alpha,\beta,l_{ij}\}$ for $i,j=0,1,\cdots,N-1$ are given under some conditions, and unknown under the other conditions. No other information is available. The node $n_{N}$ is either present or absent. If present, it is either an index node or an intermediate node. The parameters $l_{iN}$ and $l_{Nj}$ are unknown under any conditions. An empirical law $\gamma_{ij}=\Gamma_{ij}(\mbox{\boldmath{$l$}})$ is postulated in this study. Without the law, when $\mbox{\boldmath{$\theta$}}$ were unknown, the continuous parameter $\mbox{\boldmath{$\gamma$}}$ would be estimated, rather than the binary parameter $\mbox{\boldmath{$l$}}$.

\section{Method}
\label{method}

\subsection{Probability density function}
\label{probability}
The time evolution of $I_{i}(t)$, or $J_{i}(t)$ is given by Langevin equations \cite{Huf04}. They are a system of stochastic differential equations \cite{Kam07}. The ensemble of an infinite number of sample trajectories $I_{i}(t)$ is equivalent to the time dependent joint probability density function $P(\mbox{\boldmath{$I$}},t)$ of the probability variables $\mbox{\boldmath{$I$}}=(I_{0},I_{1},\cdots)$. Stochastic differential equations for $I_{i}(t)$ are converted to a partial differential equation for $P(\mbox{\boldmath{$I$}},t)$. This Fokker-Planck equation \cite{Kam07} is converted to ordinary differential equations to calculate the moments of $I_{i}$ one order after another. The first order moments $m_{i}(t|\mbox{\boldmath{$\theta$}}) = \int I_{i} P(\mbox{\boldmath{$I$}},t) {\rm d}\mbox{\boldmath{$I$}}$ are the mean of $I_{i}$ at $t$. The second order moments $v_{ij}(t|\mbox{\boldmath{$\theta$}})$ are the covariance about the mean between $I_{i}$ and $I_{j}$. The third order moments $s_{ijk}(t|\mbox{\boldmath{$\theta$}})$ are the skewness about the mean among $I_{i}$, $I_{j}$, and $I_{k}$. The fourth order moments $\kappa_{ijkl}(t|\mbox{\boldmath{$\theta$}})$ are the kurtosis about the mean among $I_{i}$, $I_{j}$, $I_{k}$, and $I_{l}$. The fifth and higher order moments are not analyzed. The formulae for the moments are listed in Appendix \ref{AppA}. The observations $I_{i}(t_{d})$ at $t_{d}$ are the initial condition to obtain the moments at $t_{d+1}$. In the following, the lower order moments refer to the mean and variance, and the higher order moments refer to the skewness and kurtosis. For small $\Delta t$, $s_{ijk}(t_{d+1}|\mbox{\boldmath{$\theta$}}), \kappa_{ijkl}(t_{d+1}|\mbox{\boldmath{$\theta$}}) \ll m_{i}(t_{d+1}|\mbox{\boldmath{$\theta$}}), v_{ij}(t_{d+1}|\mbox{\boldmath{$\theta$}})$. The probability density function is a multi-variate normal distribution in eq.(\ref{norm1}). The $i$-th element of the row vector $\mbox{\boldmath{$m$}}(t_{d+1}|\mbox{\boldmath{$\theta$}})$ is $m_{i}(t_{d+1}|\mbox{\boldmath{$\theta$}})$. The $i$-th row and $j$-th column element of the matrix $\mbox{\boldmath{$v$}}(t_{d+1}|\mbox{\boldmath{$\theta$}})$ is $v_{ij}(t_{d+1}|\mbox{\boldmath{$\theta$}})$.
\begin{eqnarray}
P(\mbox{\boldmath{$I$}},t_{d+1}) \approx P_{{\rm N}}(\mbox{\boldmath{$I$}};\mbox{\boldmath{$m$}}(t_{d+1}|\mbox{\boldmath{$\theta$}}),\mbox{\boldmath{$v$}}(t_{d+1}|\mbox{\boldmath{$\theta$}})).
\label{norm1}
\end{eqnarray}

$P(\mbox{\boldmath{$I$}},t)$ depends on time and many probability variables. Analyzing a dataset is an involved task. Time dependent conditional $z$-score for $I_{i}$ resolves the involvedness. The variables $I_{i}$ at $t_{d+1}$ are converted to the variables $z_{i}$ defined by eq.(\ref{zscore}).
\begin{eqnarray}
z_{i} = \frac{I_{i}-m_{i}^{{\rm C}}(t_{d+1}|\mbox{\boldmath{$\theta$}})}{\sqrt{v_{ii}^{{\rm C}} (t_{d+1}|\mbox{\boldmath{$\theta$}})}}.
\label{zscore}
\end{eqnarray}

In eq.(\ref{zscore}), the mean $m_{i}^{{\rm C}} (t_{d+1}|\mbox{\boldmath{$\theta$}})$ and the variance $v_{ii}^{{\rm C}} (t_{d+1}|\mbox{\boldmath{$\theta$}})$ are conditioned on the observation for the rest of the variables $\mbox{\boldmath{$I$}}_{\bar{i}} = (I_{0},\cdots,I_{i-1},I_{i+1},\cdots,I_{N-1})$ at $t_{d+1}$. Generally, given $\mbox{\boldmath{$I$}}_{\bar{i}}$, the conditional probability density function for $I_{i}$ is a uni-variate normal distribution $P_{{\rm N}}(I_{i};m_{i}^{{\rm C}},v_{ii}^{{\rm C}})$ if $\mbox{\boldmath{$I$}}$ obeys a multi-variate normal distribution $P_{{\rm N}}(\mbox{\boldmath{$I$}};\mbox{\boldmath{$m$}},\mbox{\boldmath{$v$}})$. The conditional mean $m_{i}^{{\rm C}} (t_{d+1}|\mbox{\boldmath{$\theta$}})$ is given by eq.(\ref{condmean}). It is a sum of the unconditional mean $m_{i}$ and a term dependent on the difference between the observation and the expected value $(\mbox{\boldmath{$I$}}-\mbox{\boldmath{$m$}})$ for the rest of the variables at $t_{d+1}$. The column vector $(\mbox{\boldmath{$I$}}-\mbox{\boldmath{$m$}})^{{\rm T}}$ is a transpose of a row vector $\mbox{\boldmath{$I$}}-\mbox{\boldmath{$m$}}$.
\begin{eqnarray}
m_{i}^{{\rm C}} (t_{d+1}|\mbox{\boldmath{$\theta$}}) = m_{i}(t_{d+1}|\mbox{\boldmath{$\theta$}}) + \mbox{\boldmath{$v$}}_{i\bar{i}}(t_{d+1}|\mbox{\boldmath{$\theta$}}) \mbox{\boldmath{$v$}}_{\bar{i}\bar{i}}(t_{d+1}|\mbox{\boldmath{$\theta$}})^{-1}
(\mbox{\boldmath{$I$}}_{\bar{i}}-\mbox{\boldmath{$m$}}_{\bar{i}}(t_{d+1}|\mbox{\boldmath{$\theta$}}))^{{\rm T}}.
\label{condmean}
\end{eqnarray}

The conditional variance $v_{ii}^{{\rm C}} (t_{d+1}|\mbox{\boldmath{$\theta$}})$ is a Shur compliment of $\mbox{\boldmath{$v$}}_{\bar{i}\bar{i}}$ in $\mbox{\boldmath{$v$}}$. It is given by eq.(\ref{condvar}). Because the observation reduces uncertainty, $v_{ii}^{{\rm C}}$ is smaller than $v_{ii}$ by the amount determined by the second term.
\begin{eqnarray}
v_{ii}^{{\rm C}} (t_{d+1}|\mbox{\boldmath{$\theta$}}) = v_{ii}(t_{d+1}|\mbox{\boldmath{$\theta$}}) - \mbox{\boldmath{$v$}}_{i\bar{i}}(t_{d+1}|\mbox{\boldmath{$\theta$}})
\mbox{\boldmath{$v$}}_{\bar{i}\bar{i}}(t_{d+1}|\mbox{\boldmath{$\theta$}})^{-1} \mbox{\boldmath{$v$}}_{\bar{i}i}(t_{d+1}|\mbox{\boldmath{$\theta$}}).
\label{condvar}
\end{eqnarray}

In eq.(\ref{condmean}) and (\ref{condvar}), the unconditional mean is partitioned into the $i$-th element and a row vector. The unconditional covariance is partitioned into four sub-matrices ($1\times1$, $1\times(N-1)$, $(N-1)\times1$, $(N-1)\times(N-1)$ matrices). These are given by eq.(\ref{vpart}).
\begin{eqnarray}
\mbox{\boldmath{$m$}} = (m_{i}, \mbox{\boldmath{$m$}}_{\bar{i}}), \ \mbox{\boldmath{$v$}} =
\begin{pmatrix}
v_{ii} & \mbox{\boldmath{$v$}}_{i\bar{i}} \\
\mbox{\boldmath{$v$}}_{\bar{i}i} & \mbox{\boldmath{$v$}}_{\bar{i}\bar{i}} \\
\end{pmatrix}
\label{vpart}
\end{eqnarray}

The non-uniform growth at different nodes at different times is absent in $z_{i}$. Eq.(\ref{norm1}) becomes eq.(\ref{norm2}), which is valid for the all nodes $n_{i}$ at any time $t$ as far as the approximation that $\mbox{\boldmath{$I$}}$ obeys a multi-variate normal distribution holds true.
\begin{eqnarray}
P(z_{i},t_{d+1}|\mbox{\boldmath{$I$}}_{\bar{i}}, \mbox{\boldmath{$\theta$}}) = P_{{\rm N}}(z_{i};0,1). 
\label{norm2}
\end{eqnarray}

In the above discussion, it is assumed that $\mbox{\boldmath{$\theta$}}$ is known and $I_{i}(t_{d})$ is given as a dataset. If $\mbox{\boldmath{$\theta$}}$ is unknown, the network topology and transmission parameters must be estimated from a given dataset. A well-known statistical technique for this purpose is the maximal likelihood estimation or the maximal a posteriori probability estimation \cite{Has01}. The true parameter $\mbox{\boldmath{$\theta$}}$ is substituted by the estimator $\hat{\mbox{\boldmath{$\theta$}}}$. If $\Delta J_{i}(t_{d})$ is given, instead of $I_{i}(t_{d})$, $\Delta J_{i}(t_{d})$ is converted to $I_{i}(t_{d})$ by eq.(\ref{JtoI}). The all formulae for $I_{i}(t_{d})$ are applicable then.
\begin{eqnarray}
I_{i}(t_{d}) \approx \frac{\Delta J_{i}(t_{d})}{\alpha \Delta t}.
\label{JtoI}
\end{eqnarray}

Eq.(\ref{JtoI}) is a discrete-time approximation to the stochastic differential equation. The fluctuation terms are simply discarded. The motivation to use eq.(\ref{JtoI}) is as follows. The model in this study is a hidden Markovian model in the sense that the observation $\Delta J_{i}(t_{d})$ is determined by the unobserved states of variables $I_{i}(t)$ in a Markovian stochastic process. It is known generally intractable to estimate the model with many continuous-time dependent (so-called hetero-skedastic) continuous latent variables. Such an approximation as eq.(\ref{JtoI}) is critical in the estimation. If the true value $\alpha$ in eq.(\ref{JtoI}) is unknown, it is substituted by the estimator $\hat{\alpha}$. The mathematical details for these estimation and conversion are presented in \cite{Mae10}.

\subsection{Perturbation theory}
\label{perturbation}
A perturbation theory quantifies the impact of a missing spreader node on its neighboring nodes and non-neighboring nodes. Let us investigate a simple network which consists of a missing spreader node $n_{{\rm s}}$, a neighboring node $n_{{\rm n}}$ which is connected to $n_{{\rm s}}$, and a non-neighboring node $n_{{\rm a}}$ which stays apart from $n_{{\rm s}}$. That is, $N=2$ and $n_{2}=n_{{\rm s}}$. The three nodes are connected with two links between $n_{{\rm n}}$ and $n_{{\rm a}}$, and $n_{{\rm n}}$ and $n_{{\rm s}}$. Assume that $\gamma_{{\rm na}}=\gamma_{{\rm an}}=\gamma$, $\gamma_{{\rm ns}}=\gamma_{{\rm sn}}=\gamma'$, and $\gamma_{{\rm as}}=\gamma_{{\rm sa}}=0$. The presence of $n_{{\rm s}}$ means $\gamma'>0$. Given $\mbox{\boldmath{$\theta$}}=\{\alpha,\beta,\gamma\}$, how does non-zero $\gamma'$ disturb the moments of $I_{{\rm n}}$ and $I_{{\rm a}}$?

The formulae for the moments in this network are listed in Appendix \ref{AppB}. The mean for $n_{{\rm n}}$ changes in the direction determined by the sign of $I_{{\rm n}}(t_{d})-I_{{\rm s}}(t_{d})$. The variance increases in proportion to $\gamma'$. The terms dependent on $O(\gamma'^{2})$ appear in the formulae for the skewness and kurtosis. The mean and variance of the $z$-score for $n_{{\rm n}}$ are given by eq.(\ref{nz1mo}) and (\ref{nz2mo}). The quantity $\langle z \rangle$ is the average over the observations at different times. In eq.(\ref{zscore}), the disturbance is included in the observation $I_{i}$ whose moments are $m_{i}(t_{d+1}|\mbox{\boldmath{$\theta$}},\gamma')$ and $v_{ii}(t_{d+1}|\mbox{\boldmath{$\theta$}},\gamma')$. On the other hand, nothing can be assumed about the presence of $n_{{\rm s}}$ in calculating $m_{i}^{{\rm C}}$ and $v_{ii}^{{\rm C}}$. These are the values predicted from $m_{i}(t_{d+1}|\mbox{\boldmath{$\theta$}},0)$ and $v_{ii}(t_{d+1}|\mbox{\boldmath{$\theta$}},0)$ when $\gamma'=0$, given $\mbox{\boldmath{$\theta$}} =\{\alpha,\beta,\gamma \}$. 
\begin{eqnarray}
\langle z_{{\rm n}} \rangle = -\frac{\gamma' (I_{{\rm n}}-I_{{\rm s}})}{\sqrt{(\alpha+\beta)I_{{\rm n}} + \gamma(I_{{\rm n}}+I_{{\rm a}})}} \sqrt{\Delta t}.
\label{nz1mo}
\end{eqnarray}
\begin{eqnarray}
\langle (z_{{\rm n}} - \langle z_{{\rm n}} \rangle)^{2} \rangle = 1 + \frac{\gamma' (I_{{\rm n}}+I_{{\rm s}})}{(\alpha+\beta)I_{{\rm n}} + \gamma(I_{{\rm n}}+I_{{\rm a}})}.
\label{nz2mo}
\end{eqnarray}

As $\gamma'$ increases, the difference between the probability density function of $z_{{\rm n}}$ and the standardized normal distribution becomes more significant. In contrast, the mean, variance, and skewness for $n_{{\rm a}}$ do not change at all. Interestingly, the kurtosis increases in proportion to $\gamma'$. The presence of $n_{{\rm s}}$ disturbs the fourth and higher order moments for $n_{{\rm a}}$. In terms of the $z$-score for $n_{{\rm a}}$, $\langle z_{{\rm a}} \rangle = 0$ and $\langle (z_{{\rm a}} - \langle z_{{\rm a}} \rangle)^{2} \rangle = 1$. The coupling with a spreader and non-neighboring nodes emerges at this order. But such a signal from non-zero $\gamma'$ may be too weak to detect from $n_{{\rm a}}$. The above discussion implies that the normality of the probability density function for neighboring nodes is vulnerable to the perturbation which a missing spreader node exerts, but that the impact on the normality for non-neighboring nodes is not salient. This is the basis to discriminate between the neighboring nodes and non-neighboring nodes statistically.

{\em Let us extend the theory to multiple non-neighboring nodes. The entire set of nodes is treated as a big node $n_{{\rm a}}$. The number of infectious persons at the individual nodes is roughly the same as $I_{{\rm n}}$. Assume there are $k$ such nodes. $I_{{\rm a}} \approx k I_{{\rm n}}$ holds. If $n_{{\rm s}}$ is an index node, $I_{{\rm s}} \gg I_{{\rm n}}$ holds. The variance of $z_{{\rm n}}$ in eq.(\ref{nz2mo}) is sufficiently large as a signal when $k < \gamma' I_{{\rm s}}/\gamma I_{{\rm n}}$. Any $k$ satisfy this criterion unless $\gamma'$ is extremely small. If $n_{{\rm s}}$ is an intermediate node, $I_{{\rm s}} \approx I_{{\rm n}}$ holds. The variance is large enough when $k < (2\gamma'-\gamma-\alpha-\beta)/\gamma$. Small $k$, rather than small $N$, seems a prerequisite for discrimination.}

{\em When a household or a hospital ward form a sub-population, the population of a node is very small. Most of the sub-population are likely to get infected immediately at the onset of infection. The day-by-day fluctuating growth of the number of patients, in which the trace of perturbation is left, will not be observed. The formulae in this study may not work under this condition. Moreover as an extreme, a social network model, where an individual is a node, is suitable to analyze the transmission of such diseases as Acquired Immune Deficiency Syndrome (AIDS) \cite{Pot99}. Formulating a stochastic model and perturbation theory for the social network model is beyond the scope of this study.}

\subsection{Statistical discriminator}
\label{Discriminator}

\subsubsection{Mid-body discriminator}
\label{KST}
The mid-body discriminator is founded on the Kolmogorov-Smirnov test. The Kolmogorov-Smirnov test \cite{Pre07} estimates the minimal distance between two cumulative density functions. In many applications, the test is used for a one sample hypothesis testing where the null hypothesis is that the cumulative density function drawn from a dataset empirically is the same as a given reference cumulative density function. A typical reference function is a cumulative normal distribution with a given mean and variance. The test is more sensitive to the difference in the mid-bodies of the cumulative density functions than that in the tail-ends. In other words, it is more sensitive to the difference in the lower order moments. Let $z(d)$ denote multiple observations ($d=0,1,\cdots,D-1$) for a single probability variable $z$. The empirical cumulative density function $F(z)$ is given by eq.(\ref{empirical}). The function $H(x)$ for a real argument $x$ is a Heaviside's function whose value is 1 when $x>0$, and 0 when $x<0$.
\begin{eqnarray}
F(z) = \frac{1}{D} \sum_{d=0}^{D-1} H(z-z(d)).
\label{empirical}
\end{eqnarray}

The test statistic $T$ is given by eq.(\ref{teststa}). $F_{{\rm R}}(z)$ is the reference cumulative density function which $z$ obeys in the hypothesis. This is the minimal distance between the cumulative density functions. The supremum $\sup x$ for a real variable $x$ is the least upper bound of $x$.
\begin{eqnarray}
T = \sqrt{D-1} \sup_{z} |F(z) - F_{{\rm R}}(z)|. 
\label{teststa}
\end{eqnarray}

The null hypothesis is rejected at the significance level of $a$ when $T > K_{a}$ where $K_{a} = K^{-1}(1-a)$. $K(x)$ is a cumulative density function of the Kolmogorov distribution for a probability variable $x>0$. $K_{a} = 1.38$ for $a=5\%$, and $K_{a} = 1.63$ for $a=1\%$. Eq.(\ref{empirical}) is applicable to calculating the empirical cumulative density function $F_{i}(z)$ of the $z$-score for $n_{i}$ in eq.(\ref{zscore}) from the observation of $I_{i}$ at $t=t_{d} \ (d=0,1,\cdots,D-1)$. Because of the property in eq.(\ref{norm2}), the reference cumulative density function is the cumulative standard normal distribution $F_{{\rm R}}(z) = (1+{\rm erf}(z/\sqrt{2}))/2$. The test statistic $T$ in eq.(\ref{teststa}) for $n_{i}$ becomes $T_{i}$ in eq.(\ref{teststa2}).
\begin{eqnarray}
T_{i} = \sqrt{D-1} \sup_{z} |\frac{1}{D-1} \sum_{d=0}^{D-2} H(z-z_{i}(t_{d+1})) - \frac{1+{\rm erf}(z/\sqrt{2})}{2}|. 
\label{teststa2}
\end{eqnarray}

It is not obvious which value of the threshold $K_{a}$ is the most suitable because the normality for non-neighboring nodes and non-normality for neighboring nodes are just an approximation. On the other hand, it is difficult to derive an analytical formula for an appropriate discrimination threshold as a function of the parameters and experimental conditions. Searching for the threshold $T^{{\rm *}}$ experimentally is rather practical. If $T_{i} > T^{{\rm *}}$, the mid-body discriminator determines that the perturbation from an unknown origin disturbs $I_{i}$, and consequently that $n_{i}$ is neighboring to a missing spreader node.

\subsubsection{Tail-end discriminator}
\label{CRT}
The tail-end discriminator is founded on the Chauvenet rejection test. The Chauvenet rejection test \cite{Tay96} detects an outlier in a given dataset. It was invented as a criterion to assess statistically whether particular one-dimensional numerical data are likely to be spurious or not, and is used widely in experimental physics and chemistry today. First, the mean and variance of a given dataset are calculated. The probability at which the individual datum is obtained under the calculated mean and variance is evaluated. Then, the datum is considered to be an outlier if the product of the probability and the number of data in the dataset is less than a given threshold. The threshold is 0.5 in the conventional Chauvenet rejection test. For example, the data are spurious if the probability is less than 0.05 when the number of data is 10.

The test statistic $L$ is the likelihood function \cite{Has01}, which is the conditional probability of the data as a function of the parameters. The conditional probability will be noticeably small if the data are spurious, that is, the data lies in the tail-ends. The test is sensitive to the anomaly in the higher order moments. The test statistic $L_{i}$ of the $z$-score for $n_{i}$ is defined by eq.(\ref{likelihood}).
\begin{eqnarray}
L_{i} = \frac{1}{D-1} \sum_{d=0}^{D-2} \ln P(z_{i},t_{d+1}|\mbox{\boldmath{$I$}}_{\bar{i}}, \mbox{\boldmath{$\theta$}})
= \frac{1}{D-1} \sum_{d=0}^{D-2} \ln P_{{\rm N}}(z_{i};0,1).
\label{likelihood}
\end{eqnarray}

According to the conventional Chauvenet rejection test, the discrimination threshold for the test statistic is $C = \ln 0.05 = -3$ when $N=10$. Recall the discussion on the significance of the absolute value of $K_{a}$ for the mid-body discriminator. Instead of relying on $C$ in the conventional Chauvenet rejection test, searching for an appropriate discrimination threshold $L^{{\rm *}}$ experimentally is rather appropriate. If $L_{i} < L^{{\rm *}}$, the tail-end discriminator determines that a node $n_{i}$ is a neighboring node.

\section{Experiment}
\label{experiment}

\subsection{Computationally synthesized dataset}
The performance of the discriminators is studies with a number of computationally synthesized test datasets. The test datasets are synthesized by numerical integration \cite{Klo92} of a set of Langevin equations in eq.(\ref{dI/dt}) for random network topologies and transmission parameters. The network is a Erd\"os-R\'enyi model. The probability at which a link is present between $n_{i}$ and $n_{j}$ ($l_{ij}=l_{ji}=1$) is a constant $\langle k_{i} \rangle/(N-1)$. The nodal degree of a node $n_{i}$ is given by $k_{i} = \sum_{j} l_{ij}$. The average over the all nodes is $\langle k_{i} \rangle$.

It is postulated that the total number of persons who moves from $n_{i}$ to $n_{j}$ per a unit time is proportional to $\sqrt{k_{i} k_{j}}$ if a link is present. This is valid generally for the world-wide airline transportation network \cite{Bar04}. Eq.(\ref{gammacalc}) determines $\gamma_{ij}$ as a function of $\mbox{\boldmath{$l$}}$. The fraction of persons who outgoes per a unit time is a constant $\gamma$ over the network.
\begin{eqnarray}
\gamma_{ij} = \Gamma_{ij}(\mbox{\boldmath{$l$}}) = \frac{l_{ij} \sqrt{k_{i} k_{j}}}{\sum_{j} l_{ij} \sqrt{k_{i} k_{j}}} \gamma.
\label{gammacalc}
\end{eqnarray}

It is also postulated that the initial population $P_{i}(0)=S_{i}(0)+I_{i}(0)+R_{i}(0)$ of a node $n_{i}$ is proportional to the total number of persons who outgoes from the node per a unit time \cite{Mae10}. $P_{i}(0)$ is given by eq.(\ref{nodepop}). The total population is $P=10^{6} N$ in the experiment. This is relevant in synthesizing the test datasets, but not necessary in discrimination.
\begin{eqnarray}
P_{i}(0) = \frac{\sum_{j} l_{ij} \sqrt{k_{i} k_{j}}}{\sum_{i,j} l_{ij} \sqrt{k_{i} k_{j}}} P.
\label{nodepop}
\end{eqnarray}

A receiver operating characteristic curve \cite{Faw06} is drawn to evaluate the accuracy of discrimination. In signal processing, a receiver operating characteristic curve is a plot of the true positive ratio $R_{{\rm TP}}$ on the vertical axis and the false positive ratio $R_{{\rm FP}}$ on the horizontal axis for a binary discriminator as its discrimination threshold is varied. It is generally a concave function. The discrimination threshold is $T^{{\rm *}}$ for the mid-body discriminator, and $L^{{\rm *}}$ for the tail-end discriminator. There are four possible outcomes from the discriminator. They are summarized in Table \ref{table1}. True positive and true negative are right answers, but false positive and false negative are wrong answers. The number of the true positive is $N_{{\rm TP}}$. The number of nodes is $N = N_{{\rm TP}} + N_{{\rm FN}} + N_{{\rm FP}} + N_{{\rm TN}}$. The ratios are defined by eq.(\ref{tpr}). $R_{{\rm TP}}$ is called recall alternatively, and $R_{{\rm FP}}$ is called fallout.
\begin{eqnarray}
R_{{\rm TP}} = \frac{N_{{\rm TP}}}{N_{{\rm TP}} + N_{{\rm FN}}}, \ R_{{\rm FP}} = \frac{N_{{\rm FP}}}{N_{{\rm FP}} + N_{{\rm TN}}}.
\label{tpr}
\end{eqnarray}

\begin{table}
\caption{Four possible outcomes from a discriminator.}
\begin{center}
\begin{tabular}{|c|c|c|c|}
\hline
\multicolumn{2}{|c|}{} & \multicolumn{2}{c|}{Actual value} \\
\cline{3-4}
\multicolumn{2}{|c|}{} & Neighboring & Non-neighboring \\
\hline
Discriminator & Neighboring & True positive $N_{{\rm TP}}$ & False negative $N_{{\rm FP}}$\\
\cline{2-4}
output & Non-neighboring & False positive $N_{{\rm FN}}$& True negative $N_{{\rm TN}}$\\
\hline
\end{tabular}
\end{center}
\label{table1}
\end{table}

The ratios take the value between $0$ and $1$. If the discriminator works ideally excellently, $N_{{\rm TP}} = k_{N}$, $N_{{\rm FN}} = 0$, $N_{{\rm FP}} = 0$, and $N_{{\rm TN}} = N-k_{N}$. The curve for the ideal discriminator degenerates to the upper left corner $(R_{{\rm FP}},R_{{\rm TP}})=(0,1)$. The curve for a random discriminator is a straight line $R_{{\rm TP}}=R_{{\rm FP}}$ between $(0,0)$ and $(1,1)$. The curve for a more excellent discriminator comes closer to the upper left corner. The closeness of $R_{{\rm TP}}-R_{{\rm FP}}$ to its ideal value of 1 is a scalar indicator of accuracy. It is used as an objective function to search for the optimal thresholds $L^{{\rm *}}$ and $T^{{\rm *}}$ experimentally.

\begin{figure}
\begin{center}
\includegraphics[scale=0.42,angle=-90]{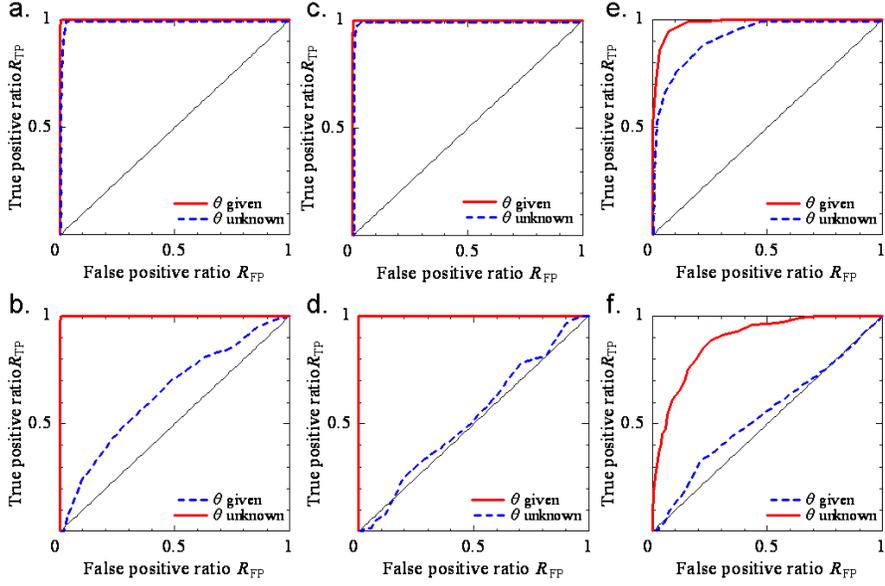}
\end{center}
\caption{Receiver operating characteristic curves of the tail-end discriminator ({\bf a}, {\bf c}, {\bf e}), and the mid-body discriminator ({\bf b}, {\bf d}, {\bf f}). The parameters are $\langle k_{i} \rangle=2$, $r=2$ ($\alpha=0.067$, $\beta=0.033$), and $\gamma=0.1$. $I_{i}(t_{d})$ for $0 \leq d \leq 99 \ (D=100)$ with $\Delta t=1$ is given as a dataset. {\bf a}, {\bf b}, $N=10$, the missing spreader node $n_{10}$ is an index node, and $I_{10}(0)=200$. {\bf c}, {\bf d}, $N=30$, $n_{30}$ is an index node, and $I_{30}(0)=200$. {\bf e}, {\bf f}, $N=10$, $n_{10}$ is not an index node but an intermediate node, and $I_{0}(0)=200$. The curves are drawn from the trials for 100 different random network topologies.}
\label{022301s}
\end{figure}

Figure \ref{022301s} {\bf a}, {\bf b} shows the receiver operating characteristic curves for $N=10$ when the missing spreader node $n_{10}$ is an index node. The tail-end discriminator works excellently both when $\mbox{\boldmath{$\theta$}}$ is given and unknown. The best performance is $R_{{\rm TP}}-R_{{\rm FP}} \approx 1$. The mid-body discriminator is the most suitable when $\mbox{\boldmath{$\theta$}}$ is given. The best performance is $R_{{\rm TP}}-R_{{\rm FP}} \approx 1$ for given $\mbox{\boldmath{$\theta$}}$, and $\approx 0.2$ for unknown $\mbox{\boldmath{$\theta$}}$. Figure \ref{022301s} {\bf c}, {\bf d} shows the curves for $N=30$ when the missing spreader node $n_{30}$ is an index node. Figure \ref{022301s} {\bf e}, {\bf f} shows the curves for $N=10$ when the missing spreader node $n_{10}$ is an intermediate node. Discrimination is not as excellent as that in {\bf a}, {\bf b}, but excellent moderately. The best performance of the tail-end discriminator is $R_{{\rm TP}}-R_{{\rm FP}} \approx 0.9$ for given $\mbox{\boldmath{$\theta$}}$, and $\approx 0.7$ for unknown $\mbox{\boldmath{$\theta$}}$. The best performance of the mid-body discriminator is $R_{{\rm TP}}-R_{{\rm FP}} \approx 0.65$ for given $\mbox{\boldmath{$\theta$}}$. The intermediate node and its neighboring nodes look alike as sinks of infected travelers indistinguishably. This is in strong contrast to the index node which is a salient source of them.

When $\mbox{\boldmath{$\theta$}}$ is unknown, the estimation searches for a network topology of $N$ nodes and transmission parameters, whose behavior bears the closest resemblance to $I_{0},I_{1},\cdots,I_{N-1}$ in the actual network topology of $N+1$ nodes. In the simple network in \ref{perturbation}, this results in $m_{i}(t_{d+1}|\hat{\mbox{\boldmath{$\theta$}}},0) \sim m_{i}(t_{d+1}|\mbox{\boldmath{$\theta$}},\gamma')$ and $v_{ij}(t_{d+1}|\hat{\mbox{\boldmath{$\theta$}}},0) \sim v_{ij}(t_{d+1}|\mbox{\boldmath{$\theta$}},\gamma')$, consequently $\langle z_{{\rm n}} \rangle \sim 0$ and $\langle (z_{{\rm n}} - \langle z_{{\rm n}} \rangle)^{2} \rangle \sim 1$, rather than eq.(\ref{nz1mo}) and (\ref{nz2mo}). This is confirmed by measuring the moments with the datasets. The measured moments are shown in Appendix \ref{AppC}. When a missing spreader node is absent, the moments for given $\mbox{\boldmath{$\theta$}}$ are almost the same as those for the estimator $\hat{\mbox{\boldmath{$\theta$}}}$. The estimator reproduces the spread accurately. The moments are $\langle m_{i} \rangle \approx 0$ and $\langle v_{ii} \rangle =1$. On the other hand, the difference in every moment between neighboring nodes and non-neighboring nodes for the estimator $\hat{\mbox{\boldmath{$\theta$}}}$ is much smaller than that for given $\mbox{\boldmath{$\theta$}}$ when the missing spreader node is an index node. The difference in $\langle m_{i} \rangle$, and that in $\langle v_{ii} \rangle$ are so small that $n_{{\rm n}}$ can not be distinguished from $n_{{\rm a}}$ by examining the lower order moments. This is the reason why the mid-body discriminator is not suitable if $\mbox{\boldmath{$\theta$}}$ is unknown. Furthermore, the difference in every moment between neighboring nodes and non-neighboring nodes is particularly small when the missing spreader node is an intermediate node. Discovering a missing intermediate node is a hard problem.

\begin{figure}
\begin{center}
\includegraphics[scale=0.42,angle=-90]{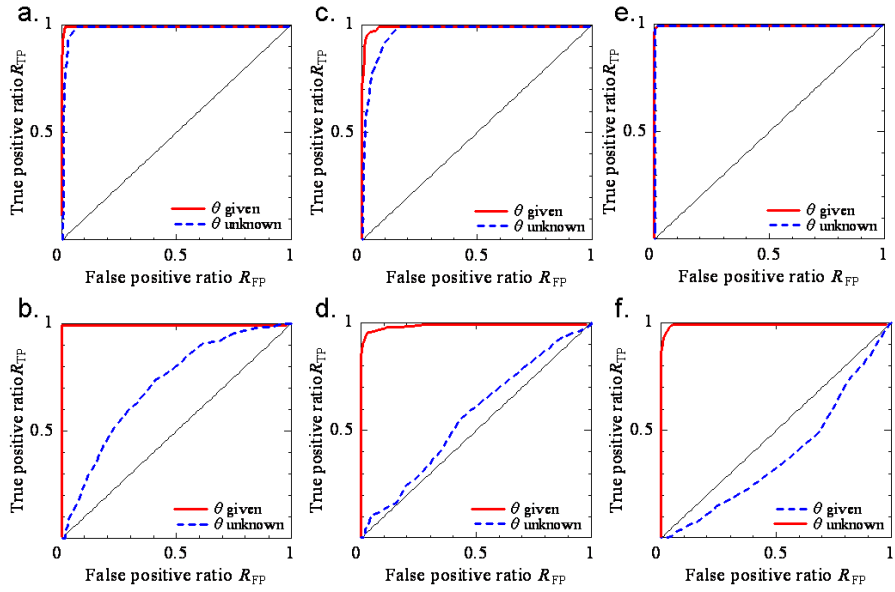}
\end{center}
\caption{Receiver operating characteristic curves of the tail-end discriminator ({\bf a}, {\bf c}, {\bf e}), and the mid-body discriminator ({\bf b}, {\bf d}, {\bf f}). {\bf a}, {\bf b}, $I_{i}(t_{d})$ for $0 \leq d \leq 32 \ (D=33)$ is given as a dataset. {\bf c}, {\bf d}, $r=8$ ($\alpha=0.089$, $\beta=0.011$). {\bf e}, {\bf f}, $\gamma=0.4$. The other experimental conditions are the same as those in Figure \ref{022301s} {\bf a}, {\bf b}.}
\label{022302s}
\end{figure}

Figure \ref{022302s} {\bf a}, {\bf b} shows the curves when the number of data is $D=33$. Accuracy is the same as that in Figure \ref{022301s} {\bf a}, {\bf b}. It does not depend on these experimental conditions. Figure \ref{022302s} {\bf c}, {\bf d} shows the curves when the reproductive ratio is $r=8$, which is four times larger than the value in Figure \ref{022301s} {\bf a}, {\bf b}. The best performance of the tail-end discriminator is $R_{{\rm TP}}-R_{{\rm FP}} \approx 0.95$ for given $\mbox{\boldmath{$\theta$}}$, and $\approx 0.85$ for unknown $\mbox{\boldmath{$\theta$}}$. The best performance of the mid-body discriminator is $R_{{\rm TP}}-R_{{\rm FP}} \approx 0.95$ for given $\mbox{\boldmath{$\theta$}}$. There is a small degradation in accuracy, but discrimination is still excellent. Figure \ref{022302s} {\bf e}, {\bf f} shows the curves when the fraction is $\gamma=0.4$, which is four times larger than the value in Figure \ref{022301s} {\bf a}, {\bf b}. The accuracy is the same as that in Figure \ref{022301s} {\bf a}, {\bf b}.

\begin{figure}
\begin{center}
\includegraphics[scale=0.42,angle=-90]{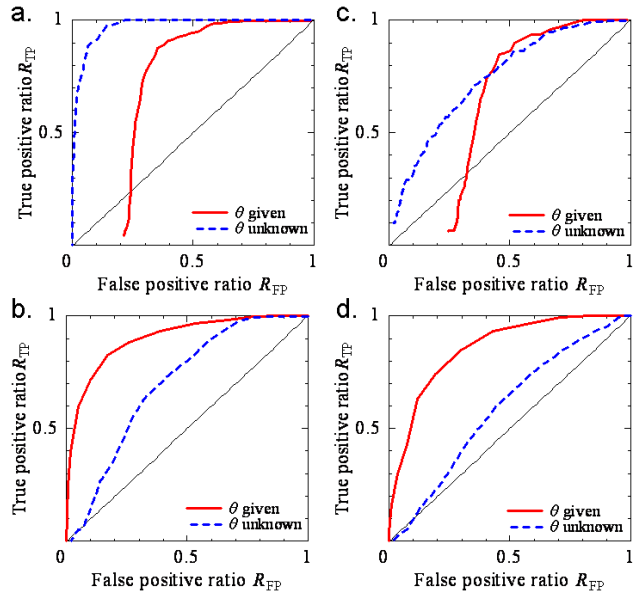}
\end{center}
\caption{Receiver operating characteristic curves of the tail-end discriminator ({\bf a}, {\bf c}), and the mid-body discriminator ({\bf b}, {\bf d}). $\Delta J_{i}(t_{d})$ for $0 \leq d \leq 99 \ (D=100)$ with $\Delta t=1$ is given as a dataset. {\bf a}, {\bf b}, the missing spreader node $n_{10}$ is an index node, and $I_{10}(0)=200$. {\bf c}, {\bf d}, $n_{10}$ is not an index node but an intermediate node, and $I_{0}(0)=200$. The experimental conditions are the same as those in Figure \ref{022301s} {\bf a}, {\bf b}, {\bf e}, {\bf f}.}
\label{022303s}
\end{figure}

Figure \ref{022303s} {\bf a}, {\bf b} shows the curves when $\Delta J_{i}(t_{d})$, instead of $I_{i}(t_{d})$, is given as a dataset. The missing spreader node $n_{10}$ is an index node. The experimental conditions are the same as those in Figure \ref{022301s} {\bf a}, {\bf b}. These two spreads are identical. Discrimination from $\Delta J_{i}(t_{d})$ is not so excellent as than from $I_{i}(t_{d})$ in Figure \ref{022301s} {\bf a}, {\bf b}. The best performance of the tail-end discriminator is $R_{{\rm TP}}-R_{{\rm FP}} \approx 0.55$ for given $\mbox{\boldmath{$\theta$}}$, and $\approx 0.85$ for unknown $\mbox{\boldmath{$\theta$}}$. The best performance of the mid-body discriminator is $R_{{\rm TP}}-R_{{\rm FP}} \approx 0.65$ for given $\mbox{\boldmath{$\theta$}}$. Figure \ref{022303s} {\bf c}, {\bf d} shows the curves under the condition where the missing spreader node $n_{10}$ is an intermediate node. The best performance of the tail-end discriminator is $R_{{\rm TP}}-R_{{\rm FP}} \approx 0.4$ both for given and unknown $\mbox{\boldmath{$\theta$}}$. The best performance of the mid-body discriminator is $R_{{\rm TP}}-R_{{\rm FP}} \approx 0.6$ for given $\mbox{\boldmath{$\theta$}}$. The performance of the mid-body discriminator does not change much between {\bf a}, {\bf b} and {\bf c}, {\bf d}.

Somehow these tendencies are contrary to the results so far. Recall that the fluctuation term in eq.(\ref{dJ/dt}) is discarded in the approximation of eq.(\ref{JtoI}). The term is more complex than a mere Gaussian white noise because its amplitude includes $\sqrt{I_{i}(t)}$, and the ensemble of sample trajectories $I_{i}(t)$ does not obey a multi-variate normal distribution in eq.(\ref{norm1}) for large $t$. Eq.(\ref{JtoI}) is correct on the average, but the conversion of the higher order moments from $\Delta J_{i}(t_{d})$ to $I_{i}(t_{d})$ is inaccurately distortive. It is the reason why the tail-end discriminator is only moderately excellent, and the shape of the curve looks awkward when $\mbox{\boldmath{$\theta$}}$ is given. On the other hand, when $\mbox{\boldmath{$\theta$}}$ is unknown, the estimator $\hat{\mbox{\boldmath{$\theta$}}}$ may be able to make up for the distortion in the higher order moments. The tail-end discriminator with the estimator $\hat{\mbox{\boldmath{$\theta$}}}$ outperforms the discriminators for given $\mbox{\boldmath{$\theta$}}$. Even if the true parameters are given, adjusting the parameters is rather indispensable for excellent discrimination from $\Delta J_{i}(t_{d})$. Substantial improvement of eq.(\ref{JtoI}) is beyond the scope of this study and for future challenge.

\begin{figure}
\begin{center}
\includegraphics[scale=0.42,angle=-90]{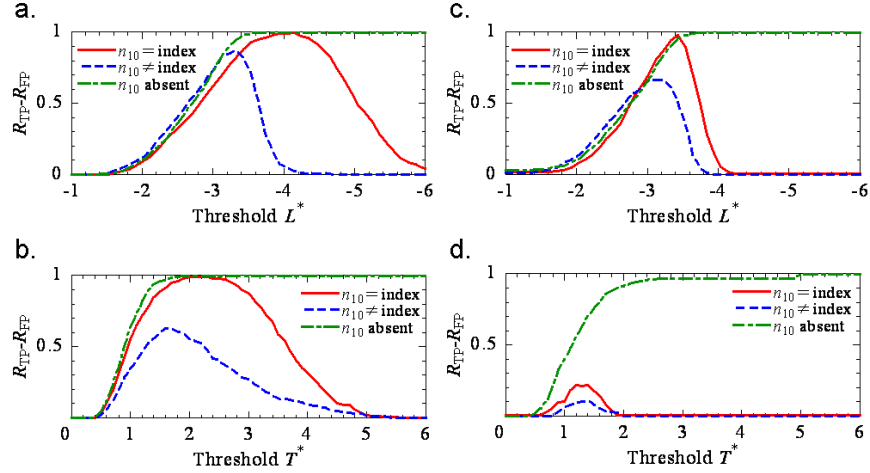}
\end{center}
\caption{$R_{{\rm TP}}-R_{{\rm FP}}$ of the tail-end discriminator ({\bf a}, {\bf c}), and the mid-body discriminator ({\bf b}, {\bf d}) as a function of the discrimination thresholds. {\bf a}, {\bf b}, $\mbox{\boldmath{$\theta$}}$ is given. {\bf c}, {\bf d}, $\mbox{\boldmath{$\theta$}}$ is unknown and estimated from the dataset. The missing spreader node $n_{10}$ is either an index node, an intermediate node, or absent. The experimental conditions are the same as those in Figure \ref{022301s} {\bf a}, {\bf b}, {\bf e}, {\bf f}.}
\label{022304s}
\end{figure}

Figure \ref{022304s} {\bf a}, {\bf b} shows $R_{{\rm TP}}-R_{{\rm FP}}$ as a function of the discrimination threshold $L^{{\rm *}}$ or $T^{{\rm *}}$ when $\mbox{\boldmath{$\theta$}}$ is given. The rising edges of the three curves coincide with each other. $L^{{\rm *}}=-3$ to $-3.5$, and $T^{{\rm *}}=1.5$ to $2$ are reasonable if there is no prior information on the presence or absence of a missing spreader node. If the assumption is grounded well that an index node is missing, there is a distinct optimal threshold $L^{{\rm *}}=-4$, and $T^{{\rm *}}=2.1$. Figure \ref{022304s} {\bf c}, {\bf d} shows $R_{{\rm TP}}-R_{{\rm FP}}$ when $\mbox{\boldmath{$\theta$}}$ is unknown. Setting $L^{{\rm *}}=-3$ to $-3.5$ is reasonable. The position of the falling edges of the curves in {\bf c}, {\bf d} are different from that in {\bf a}, {\bf b}. There is a distinct optimal threshold $L^{{\rm *}}=-3.4$ for a missing index node.

\begin{figure}
\begin{center}
\includegraphics[scale=0.42,angle=-90]{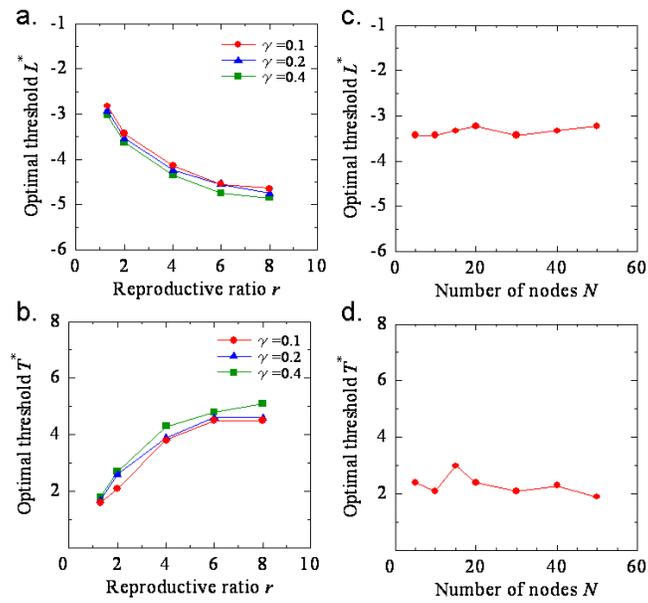}
\end{center}
\caption{Optimal discrimination thresholds of the tail-end discriminator for unknown $\mbox{\boldmath{$\theta$}}$ ({\bf a}, {\bf c}), and the mid-body discriminator for given $\mbox{\boldmath{$\theta$}}$ ({\bf b}, {\bf d}). {\bf a}, {\bf b}, as a function of $r$ and $\gamma$. {\bf c}, {\bf d}, as a function of $N$. The experimental conditions are the same as those in Figure \ref{022301s} {\bf a}, {\bf b}.}
\label{022305s}
\end{figure}

Figure \ref{022305s} {\bf a}, {\bf b} shows the optimal thresholds $L^{{\rm *}}$ and $T^{{\rm *}}$ as a function of the transmission parameters $r$ ($\alpha$ and $\beta$) and $\gamma$. Figure \ref{022305s} {\bf c}, {\bf d} shows the optimal thresholds $L^{{\rm *}}$ and $T^{{\rm *}}$ as a function of the number of nodes $N$. The optimal thresholds depend on $r$. This means that it is practical to find the optimal threshold experimentally, which is suitable for individual conditions. The dependence on $\gamma$ and $N$ is so small that the dependence can be ignored.

\subsection{WHO dataset}
\subsubsection{SARS outbreak in 2003}
SARS is a respiratory disease in humans caused by the SARS corona-virus. The epidemic of SARS appears to have started in Guangdong of south China in November 2002. SARS spread from the Guangdong to Hong Kong in early 2003 \cite{Lip03}, and eventually nearly 40 countries around the world by July \cite{Ril03}. WHO archives the cumulative number of reported probable cases of SARS\footnote{World Health Organization, Cumulative number of reported probable cases of SARS, http://www.who.int/csr/sars/country/en/index.html (2003).}. The dataset in the archive had been updated every day. It is a collection of time sequence data $J_{i}(t_{d})$ with $\Delta t=1$ day.

The target geographical regions in this study are those where five or more cases had been reported in a month since March 17. The number of data is $D=31$. They are Canada (CAN), France (FRA), United Kingdom (GBR), Germany (DEU), Hong Kong (HKG), Malaysia (MAS), Taiwan (ROC), Singapore (SIN), Thailand (THA), United States (USA), and Vietnam (VIE). Mainland China is not included because no data are available in some periods, and no data outside of Guangdong is reported in other periods. The total cumulative number of cases increased 10 times from $J(t_{0})=\sum_{i} J_{i}(t_{0}) = 165$ to $J(t_{30})=1,846$ in these $N=11$ regions. The fluctuation in the dataset originates in the observational noise partly, which arises from inaccurate diagnosis and the irregular delays in reporting. The dataset is smoothed with a moving average filter \cite{Wal10}. The window size here is $W=3$ ($\approx 0.1D$). It is postulated that the empirical law in eq.(\ref{gammacalc}) holds true.

Table \ref{SARS} shows the calculated test statistic $L_{i}$ for the mid-body discriminator, $T_{i}$ for the tail-end discriminator, and the mean $m_{i}$, variance $v_{ii}$, skewness $s_{iii}$, and kurtosis $\kappa_{iiii}$ of $z_{i}$ for the 11 regions. The lower order moments averaged over the all regions are $\langle m_{i} \rangle=-1.0$ and $\langle v_{ii} \rangle =3.2$. The entire dataset is disturbed. The lower order moments tend to be anomalous for United States, Canada, Singapore, and Taiwan, while the higher order moments are anomalously large for Malaysia and Vietnam. The estimated values of the parameters are $\hat{\alpha}=0.18$, $\hat{\beta}=0.13$ ($r=1.4$), and $\hat{\gamma}=0.13$. The optimal threshold $L^{{\rm *}}=-3.6$ is obtained in the experiment for this condition. Perturbation is discovered in the time sequence data for Hong Kong, United States, Canada, Singapore, Taiwan, Malaysia, and Vietnam. The data for Hong Kong is disturbed extremely strongly. The variance and kurtosis are anomalous particularly.

\begin{table}
\caption{Test statistic $L$ for the mid-body discriminator, $T$ for the tail-end discriminator, and the mean $m_{i}$, variance $v_{ii}$, skewness $s_{iii}$, and kurtosis $\kappa_{iiii}$ of $z_{i}$ for the 11 regions where a local outbreak is reported in the early growth phase of the SARS outbreak.}
\begin{center}
\begin{tabular}{|c|c|c|c|c|c|c|}
\hline
Region & $L$ & $T$ & $m_{i}$ & $v_{ii}$ & $s_{iii}$ & $\kappa_{iiii}$ \\
\hline
HKG & -41.3 & 2.1 & 0.67 & 8.7 & 0.39 & 5.1 \\
\hline
USA & -16.1 & 2.7 & -2.0 & 4.8 & 0.11 & -0.15 \\
\hline
CAN & -15.2 & 3.1 & -1.7 & 4.8 & 0.45 & 0.78 \\
\hline
SIN & -8.5 & 1.6 & -0.77 & 3.4 & -0.41 & 0.24 \\
\hline
ROC & -8.1 & 2.7 & -1.7 & 3.2 & 0.36 & -0.16 \\
\hline
MAS & -4.5 & 3.1 & -1.5 & 2.1 & 1.2 & 3.8 \\
\hline
VIE & -3.6 & 1.5 & -0.89 & 2.4 & -2.0 & 6.4 \\
\hline
GER & -2.4 & 2.1 & -0.99 & 1.4 & 0.29 & -0.81 \\
\hline
FRA & -2.4 & 2.0 & -0.85 & 1.5 & 0.70 & 0.47 \\
\hline
THI & -2.3 & 1.8 & -0.76 & 1.7 & 0.32 & 0.12 \\
\hline
GBR & -1.4 & 1.4 & -0.42 & 1.5 & 0.78 & 1.6 \\
\hline
Average & -9.6 & 2.2 & -1.0 & 3.2 & 0.20 & 1.6 \\
\hline
\end{tabular}
\end{center}
\label{SARS}
\end{table}

Recall that Mainland China is not included in the dataset. The actual cumulative number of cases in Guangdong alone exceeded that in Hong Kong in the period. Guangdong was a spreader which had not been known to the rest of the world in the early growth phase. This is illustrated by an example of the uncovered chains of transmission. Two of the index patients in Toronto in Canada and another three of the index patients in United States stayed a hotel in Hong Kong, where a Chinese nephrologist, who had treated many patients in Guangzhou in Guangdong and become infected, was staying in late February\footnote{SARS Expert Committee (Hong Kong), SARS in Hong Kong: from experience to action, http://www.sars-expertcom.gov.hk/english/reports/reports/reports\_fullrpt.html (2003).}. This implies that China was influential potentially in infecting Hong Kong, United States, and Canada. The large amount of passenger traffic between south China and Southeast Asian countries is supposed to affect the spread in these countries. On the other hand, the disturbance is not evident in the data for European countries. It is highly probable that the discovered signal of perturbation originates in the transmission between China, a missing spreader node, and north Pacific Rim countries.

\subsubsection{Flu pandemic in 2009}
The flu pandemic in 2009 was a global outbreak of a new strain of the H1N1 swine influenza A virus. The virus appeared in Veracruz in southeast Mexico, in April 2009. The pandemic spread to United States and Canada immediately, and then to the South American countries, West European countries, and Pacific Rim countries. It began to decline in November. WHO archives the cumulative number of the reported laboratory-confirmed cases of the flu pandemic\footnote{World Health Organization, Situation updates - Pandemic (H1N1) 2009, http://www.who.int/csr/disease/swineflu/updates/en/index.html (2010).}. The dataset in the archive had been updated every day. It is a collection of time sequence data $J_{i}(t_{d})$ with $\Delta t=1$ day.

The target geographical regions in this study are those where five or more cases had been reported in about three weeks since April 28. The number of data is $D=25$. They are Australia (AUS), Belgium (BEL), Brazil (BRA), Canada (CAN), Chile (CHL), China (CHN), Colombia (COL), Costa Rica (CRI), Ecuador (ECU), El Salvador (SLV), France (FRA), Germany (DEU), Israel (ISR), Italy (ITA), Japan (JPN), Mexico (MEX), New Zealand (NZL), Panama (PAN), Peru (PER), Spain (ESP), United Kingdom (GBR), and United States (USA). The cumulative number of cases increased 100 times from $J({t_{0}})=105$ to $J(t_{24})=11,129$ in these $N=22$ regions. The dataset is smoothed with a moving average filter whose window size is $W=3$. It is postulated that the law in eq.(\ref{gammacalc}) holds true for the pandemic.

Table \ref{H1N1} shows the calculated test statistics, the mean, variance, skewness, and kurtosis of $z_{i}$ for the 22 regions. The lower order moments averaged over the all regions are $\langle m_{i} \rangle=-0.11$ and $\langle v_{ii} \rangle =1.3$. The entire dataset is not disturbed. The lower order moments tend to be anomalous for United States and Mexico, while the higher order moments are anomalously large for Canada. Irregularly, Japan has large absolute values of both lower and higher order moments. The higher order moments for South American countries tend to be large while those for West European countries are small. This may be related to the seasonality for the tropics and north hemisphere. The estimated values of the parameters are $\hat{\alpha}=0.88$, $\hat{\beta}=0.77$ ($r=1.2$), and $\hat{\gamma}=0.29$. The optimal threshold $L^{{\rm *}}=-4.2$ is obtained in the experiment for this condition. Perturbation is discovered in the time sequence data for United States, Mexico, Canada, and Japan. The ratio of this number is $4/N=0.18$, which is remarkably smaller than $7/N=0.64$ for the SARS outbreak. In addition, the absolute value of $L$ is much less anomalous than that for the SARS outbreak. In most regions, $L$ is close to 0 while $L$ is less than -2 for the SARS outbreak. The disturbance from an unknown origin of perturbation is small and localized. Then, what happened to the spread in United States, Mexico, Canada, and Japan?

\begin{table}
\caption{Test statistics $L$, $T$, and the mean $m_{i}$, variance $v_{ii}$, skewness $s_{iii}$, and kurtosis $\kappa_{iiii}$ of $z_{i}$ for the 22 regions where a local outbreak is reported in the early growth phase of the flu pandemic.}
\begin{center}
\begin{tabular}{|c|c|c|c|c|c|c|}
\hline
Region & $L$ & $T$ & $m_{i}$ & $v_{ii}$ & $s_{iii}$ & $\kappa_{iiii}$ \\
\hline
USA & -17.2 & 2.4 & 0.07 & 5.2 & -0.09 & -0.55 \\
\hline
MEX & -16.9 & 1.8 & -1.7 & 4.8 & -0.31 & -0.10 \\
\hline
CAN & -8.4 & 0.84 & 0.02 & 3.4 & 2.6 & 9.6 \\
\hline
JPN & -7.3 & 3.3 & -1.5 & 2.8 & 2.3 & 5.3 \\
\hline
GBR & -2.4 & 1.1 & -0.43 & 0.74 & 0.48 & -0.56 \\
\hline
ESP & -2.2 & 1.0 & -0.05 & 0.80 & 1.4 & 2.4 \\
\hline
PAN & -1.9 & 1.1 & -0.12 & 0.92 & 1.5 & 2.2 \\
\hline
CRI & -1.6 & 1.5 & -0.15 & 1.1 & 3.0 & 9.6 \\
\hline
FRA & -1.2 & 0.93 & -0.02 & 0.67 & 1.6 & 3.6 \\
\hline
DEU & -1.0 & 1.3 & 0.21 & 0.75 & 2.0 & 3.4 \\
\hline
COL & -0.77 & 1.3 & -0.02 & 0.43 & 1.0 & 0.72 \\
\hline
ITA & -0.71 & 1.5 & 0.17 & 0.55 & 1.0 & 0.54 \\
\hline
CHN & -0.68 & 1.6 & -0.02 & 0.25 & 0.88 & 0.55 \\
\hline
CHL & -0.65 & 1.7 & 0.30 & 0.84 & 2.6 & 7.1 \\
\hline
SLV & -0.61 & 1.2 & -0.01 & 0.54 & 1.3 & 1.1 \\
\hline
NZL & -0.58 & 1.4 & 0.02 & 0.48 & 2.2 & 5.4 \\
\hline
BRA & -0.52 & 1.4 & -0.03 & 0.60 & 3.2 & 11.6 \\
\hline
ECU & -0.51 & 1.6 & 0.16 & 0.89 & 3.8 & 14.4 \\
\hline
PER & -0.35 & 1.9 & 0.15 & 0.51 & 2.3 & 6.1 \\
\hline
AUS & -0.27 & 1.7 & 0.27 & 0.64 & 1.9 & 2.4 \\
\hline
BEL & -0.24 & 1.8 & 0.14 & 0.47 & 2.1 & 3.6 \\
\hline
ISR & -0.24 & 1.8 & 0.08 & 0.34 & 0.81 & 0.04 \\
\hline
Average & -3.0 & 1.6 & -0.11 & 1.3 & 1.7 & 4.0 \\
\hline
\end{tabular}
\end{center}
\label{H1N1}
\end{table}

The national transportation statistics of United States\footnote{Research and Innovative Technology Administration, Bureau of Transportation Statistics, USA, http://www.bts.gov/publications/national\_transportation\_statistics (2010).} reports that most passengers cross the borders between United States and Mexico, and United States and Canada by land. The annual number of passengers from Mexico to United States by land is 24 times larger than that by air in 2009. The number from Canada to United States by land is 7 times larger than that by air in 2005. The meta-population network model relies on eq.(\ref{gammacalc}) which is known valid for the world-wide air transportation. The actual heavy land transportation lets the probability of movement across the border deviate from the value estimated in the model. This is supported by the result that the fraction $\hat{\gamma}$ is twice as large as that for the SARS outbreak. The deviation imposes an impact on the speed of the spread at such cities near the border as San Ysidro CA, then inland cities, and finally such cities near another border as Buffalo-Niagara Falls NY \cite{Bal09}. The discovered signal of perturbation for United States, Mexico, and Canada could originate in a localized exception to the empirical law for the probability of movement, rather than a missing spreader node.

The onset of infection in Japan started suddenly on May 9 after many travelers had returned from North America during the Japan's two week long festive break. Flu spread explosively among high school students in Osaka and Hyogo. The effective reproductive ratio reached the peak value around May 14 \cite{Nis09}. The ratio is significantly higher than 1.4 to 1.6 from an epidemiological analysis, or 1.2 from a genetic analysis, in Mexico and other countries \cite{Fra09}. School closure started on May 17 just after the secondary transmission had been confirmed and announced officially. The effective reproductive ratio declined below 1. The reproductive ratio in Japan rose and fell sharply in 10 days in early May. The precondition of the SIR compartment model is that the values of the transmission parameters do not depend on time and sub-populations. The discovered signal of perturbation for Japan could originate in an accidental localized anomaly on the probability of infection, rather than a missing spreader node.

The above results demonstrate the potential capability of the discriminator in discovering unknown origins of perturbation which violate the preconditions of the mathematical models. On the other hand, the signal discovered by the discriminators satisfies the sufficient condition to ascertain the presence of a missing spreader node incompletely, although it satisfies the necessary conditions. Prior knowledge on the demographic and socioeconomic nature of individual regions makes up for the incompleteness. {\em This is similar to the false positives where the signal identified an early warning for a catastrophe originates in wide classes of impending transitions because the underlying mechanism is understood incompletely \cite{Sch09}. It is also of interest to see if the discovered location of a missing node satisfies the criterion for a strongly influential spreader in the core of the network \cite{Kit10}.} Is it possible to distinguish the presence of a missing spreader node from an anomaly on the parameters mathematically? The solution will be worked out by examining how possible origins of perturbation appear in respective order moments of the disturbed time sequence data, and by finding a suitable backward mapping from the pattern of disturbance to the possible origins. The criterion for discrimination may be a complex function of the preconditions of a mathematical model, experimental conditions, and given datasets, rather than a simple scalar threshold of a test statistic in this study. These issues are for future challenges.

\section{Conclusion}
This study poses an intriguing problem which few studies have addressed before. The problem is a node discovery problem for the spread of an infectious disease. Two statistical discriminators are invented to solve the problem.

The performance of the discriminators is studied with the computationally synthesized datasets. The findings are as follows. The tail-end discriminator is excellent both when the parameters is given and unknown. The mid-body discriminator is the most suitable when the parameters are given. Discovering a missing intermediate node is more difficult than a missing index node, but possible. The performance depends neither on the speed of the spread, on the size of the network, nor on the number of data much. If the cumulative number of new cases is given as a dataset, discrimination is moderately excellent, and interestingly, the tail-end discriminator works more excellently when the parameters are unknown than when they are given. The optimal thresholds for discrimination depend solely on the reproductive ratio (the probability of infection and recovery).

The WHO dataset on the SARS outbreak and flu pandemic are analyzed with the discriminators. The findings are as follows. The entire dataset on the SARS outbreak is disturbed. The signal of perturbation from an unknown origin is discovered in the time sequence data for Hong Kong, United States, Canada, Singapore, Taiwan, Malaysia, and Vietnam. The data for Hong Kong is disturbed extremely strongly. It is highly probable that the discovered signal originates in the transmission between China, a missing spreader node, and north Pacific Rim countries. In the flu pandemic, the signal of perturbation is discovered in the time sequence data for United States, Mexico, Canada, and Japan. The ratio of the number of these regions is much smaller than that for the SARS outbreak. The test statistics are much less anomalous than those for the SARS outbreak. The disturbance in the dataset is small and localized. The signal for United States, Mexico, and Canada could originate in the heavy land transportation which is an exception to the empirical law for the probability of movement. The signal for Japan could originate in an accidentally sharp rise and fall of the probability of infection.

Two advanced problems still remain unsolved. One is raising the accuracy of discrimination from the cumulative number of new cases as well as from the number of cases. The other is deriving the criteria to distinguish the presence of a missing spreader node from an anomaly on the parameters. Solving them is a milestone to the theory on the stochastic reaction-diffusion process where hidden external or internal entities are at work. These issues are for future challenges.

\appendix

\section{Time evolution of moment}
\label{AppA}
The time evolution of $I_{i}(t)$ is given by the Langevin equations in eq.(\ref{dI/dt}). The fluctuation terms $\xi^{[\gamma]}(t)$, $\xi^{[\alpha]}(t)$, and $\xi^{[\beta]}(t)$ are Gaussian white noises. Their explicit functional forms are unknown. The equations are applicable to any nodes including missing nodes.
\begin{eqnarray}
\frac{{\rm d} I_{i}(t)}{{\rm d} t} &=& \frac{\alpha S_{i}(t) I_{i}(t)}{S_{i}(t)+I_{i}(t)+R_{i}(t)} - \beta I_{i}(t) 
+ \sum_{j} \gamma_{ji} I_{j}(t) - \sum_{j} \gamma_{ij} I_{i}(t) \nonumber \\
&+& \sqrt{\frac{\alpha S_{i}(t) I_{i}(t)}{S_{i}(t)+I_{i}(t)+R_{i}(t)}} \xi^{[\alpha]}_{i}(t) 
- \sqrt{\beta I_{i}(t)} \xi^{[\beta]}_{i}(t) \nonumber \\
&+& \sum_{j} \sqrt{\gamma_{ji} I_{j}(t)} \xi^{[\gamma]}_{ji}(t) - \sum_{j} \sqrt{\gamma_{ij} I_{i}(t)} \xi^{[\gamma]}_{ij}(t).
\label{dI/dt}
\end{eqnarray}

In most cases, the outbreak is contained before the spread reaches equilibrium. In the early growth phase of the outbreak, $I_{i} \ll S_{i}$ and $R_{i} \ll S_{i}$ hold true. Eq.(\ref{dI/dt}) becomes eq.(\ref{apdI/dt}) \cite{Mae10}.
\begin{eqnarray}
\frac{{\rm d} I_{i}(t)}{{\rm d} t} &=& \alpha I_{i}(t) - \beta I_{i}(t) 
+ \sum_{j} \gamma_{ji} I_{j}(t) - \sum_{j} \gamma_{ij} I_{i}(t) \nonumber \\
&+& 
\sqrt{\alpha I_{i}(t)} \xi^{[\alpha]}_{i}(t) - \sqrt{\beta I_{i}(t)} \xi^{[\beta]}_{i}(t) \nonumber \\
&+& \sum_{j} \sqrt{\gamma_{ji} I_{j}(t)} \xi^{[\gamma]}_{ji}(t) - \sum_{j} \sqrt{\gamma_{ij} I_{i}(t)} \xi^{[\gamma]}_{ij}(t).
\label{apdI/dt}
\end{eqnarray}

The time evolution of $J_{i}(t)$ is given by eq.(\ref{dJ/dt}).
\begin{eqnarray}
\frac{{\rm d} J_{i}(t)}{{\rm d} t} = \alpha I_{i}(t) + \sqrt{\alpha I_{i}(t)} \xi^{[\alpha]}_{i}(t).
\label{dJ/dt}
\end{eqnarray}

Eq.(\ref{dI/dt}) is equivalent to the Fokker-Planck equation in eq.(\ref{FokkerPlanck}).
\begin{eqnarray}
\frac{\partial P(\mbox{\boldmath{$I$}},t)}{\partial t} = -\sum_{i} \frac{\partial}{\partial I_{i}} (\sum_{p} a_{ip} I_{p}) P(\mbox{\boldmath{$I$}},t) + \frac{1}{2} \sum_{i,j} \frac{\partial^{2}}{\partial I_{i} \partial I_{j}} (\sum_{p} b_{ijp} I_{p}) P(\mbox{\boldmath{$I$}},t).
\label{FokkerPlanck}
\end{eqnarray}

The coefficients $a_{ip}$ and $b_{ijp}$ in eq.(\ref{FokkerPlanck}) are given by eq.(\ref{aip}) and (\ref{bijp}).
\begin{eqnarray}
a_{ip} = (\alpha-\beta-\sum_{j'} \gamma_{ij'})\delta_{ip} + \gamma_{pi}.
\label{aip}
\end{eqnarray}
\begin{eqnarray}
b_{ijp} = \{(\alpha+\beta+\sum_{j'} \gamma_{ij'}) \delta_{ip}+\gamma_{pi} \} \delta_{ij} - \gamma_{ij} \delta_{ip} - \gamma_{ji} \delta_{jp}.
\label{bijp}
\end{eqnarray}

The moments satisfy eq.(\ref{dm/dt}) through (\ref{dk/dt}). The explicit functional forms of lower order moments are necessary to obtain higher order moments.
\begin{eqnarray}
\frac{{\rm d} m_{i}(t|\mbox{\boldmath{$\theta$}})}{{\rm d}t} = \sum_{p} a_{ip} m_{p}(t|\mbox{\boldmath{$\theta$}}). 
\label{dm/dt}
\end{eqnarray}
\begin{eqnarray}
\frac{{\rm d} v_{ij}(t|\mbox{\boldmath{$\theta$}})}{{\rm d}t} = \sum_{p} ( a_{ip} v_{pj}(t|\mbox{\boldmath{$\theta$}}) + a_{jp} v_{pi}(t|\mbox{\boldmath{$\theta$}}) ) + \sum_{p} b_{ijp} m_{p}(t|\mbox{\boldmath{$\theta$}}). 
\label{dv/dt}
\end{eqnarray}
\begin{eqnarray}
\frac{{\rm d} s_{ijk}(t|\mbox{\boldmath{$\theta$}})}{{\rm d}t} &=& \sum_{p} ( a_{ip} s_{pjk}(t|\mbox{\boldmath{$\theta$}}) + a_{jp} s_{pik}(t|\mbox{\boldmath{$\theta$}}) + a_{kp} s_{pij}(t|\mbox{\boldmath{$\theta$}}) ) \nonumber \\
&+& \sum_{p} ( b_{ijp} v_{pk}(t|\mbox{\boldmath{$\theta$}}) + b_{ikp} v_{pj}(t|\mbox{\boldmath{$\theta$}}) + b_{jkp} v_{pi}(t|\mbox{\boldmath{$\theta$}}) ).
\label{ds/dt}
\end{eqnarray}
\begin{eqnarray}
\frac{{\rm d} \kappa_{ijkl}(t|\mbox{\boldmath{$\theta$}})}{{\rm d}t} &=& \sum_{p} ( a_{ip} \kappa_{pjkl}(t|\mbox{\boldmath{$\theta$}}) + a_{jp} \kappa_{pikl}(t|\mbox{\boldmath{$\theta$}}) + a_{kp} \kappa_{pijl}(t|\mbox{\boldmath{$\theta$}}) + a_{lp} \kappa_{pijk}(t|\mbox{\boldmath{$\theta$}}) ) \nonumber \\
&+& \sum_{p} ( b_{ijp} s_{pkl}(t|\mbox{\boldmath{$\theta$}}) + b_{ikp} s_{pjl}(t|\mbox{\boldmath{$\theta$}}) + b_{ilp} s_{pjk}(t|\mbox{\boldmath{$\theta$}}) \nonumber \\
&+& b_{jkp} s_{pil}(t|\mbox{\boldmath{$\theta$}}) + b_{jlp} s_{pik}(t|\mbox{\boldmath{$\theta$}}) + b_{klp} s_{pij}(t|\mbox{\boldmath{$\theta$}}) ). 
 \label{dk/dt}
\end{eqnarray}

The solutions of eq.(\ref{dm/dt}) through (\ref{dk/dt}) are given by eq.(\ref{m(deltat)}) through (\ref{k(deltat)}) when $\Delta t$ is small.
\begin{eqnarray}
m_{i}(t_{d+1}|\mbox{\boldmath{$\theta$}}) = I_{i}(t_{d}) + \sum_{p} a_{ip} I_{p}(t_{d}) \Delta t + O(\Delta t^{2}).
\label{m(deltat)}
\end{eqnarray}
\begin{eqnarray}
v_{ij}(t_{d+1}|\mbox{\boldmath{$\theta$}}) = \sum_{p} b_{ijp} I_{p}(t_{d}) \Delta t + O(\Delta t^{2}).
\label{v(deltat)}
\end{eqnarray}
\begin{eqnarray}
s_{ijk}(t_{d+1}|\mbox{\boldmath{$\theta$}}) = \frac{1}{2} \sum_{p,q} ( b_{ijp} b_{pkq} + b_{ikp} b_{pjq} + b_{jkp} b_{piq} ) I_{q}(t_{d}) \Delta t^{2} + O(\Delta t^{3}). 
\label{s(deltat)}
\end{eqnarray}
\begin{eqnarray}
\kappa_{ijkl}(t_{d+1}|\mbox{\boldmath{$\theta$}}) &=& \frac{1}{6} \sum_{p,q,r} \{ b_{ijp} (b_{pkq} b_{qlr} + b_{plq} b_{qkr} + b_{klq} b_{qpr}) 
+ b_{ikp} (b_{pjq} b_{qlr} + b_{plq} b_{qjr} + b_{jlq} b_{qpr}) \nonumber \\
&+& b_{ilp} (b_{pjq} b_{qkr} + b_{pkq} b_{qjr} + b_{jkq} b_{qpr}) 
+ b_{jkp} (b_{piq} b_{qlr} + b_{plq} b_{qir} + b_{ilq} b_{qpr}) \nonumber \\
&+& b_{jlp} (b_{piq} b_{qkr} + b_{pkq} b_{qir} + b_{ikq} b_{qpr}) 
+ b_{klp} (b_{piq} b_{qjr} + b_{pjq} b_{qir} + b_{ijq} b_{qpr}) \} I_{r}(t_{d}) \Delta t^{3} \nonumber \\
&+& O(\Delta t^{4}). 
\label{k(deltat)}
\end{eqnarray}

\section{Disturbed moment}
\label{AppB}
The diagonal elements of the moments are derived for the simple network in \ref{perturbation} which consists of $n_{{\rm n}}$, $n_{{\rm a}}$, and $n_{{\rm s}}$. Eq.(\ref{m(deltat)}) through (\ref{k(deltat)}) become eq.(\ref{mn}) through (\ref{knnnn}) for $n_{{\rm n}}$.
\begin{eqnarray}
m_{{\rm n}}(t_{d+1}|\mbox{\boldmath{$\theta$}},\gamma') &\approx& I_{{\rm n}}(t_{d}) + \{(\alpha-\beta)I_{{\rm n}}(t_{d}) - \gamma(I_{{\rm n}}(t_{d})-I_{{\rm a}}(t_{d}))\} \Delta t \nonumber \\
&-& \gamma'(I_{{\rm n}}(t_{d})-I_{{\rm s}}(t_{d})) \Delta t.
\label{mn}
\end{eqnarray}
\begin{eqnarray}
v_{{\rm nn}}(t_{d+1}|\mbox{\boldmath{$\theta$}},\gamma') &\approx& \{(\alpha+\beta)I_{{\rm n}}(t_{d}) + \gamma(I_{{\rm n}}(t_{d})+I_{{\rm a}}(t_{d}))\} \Delta t \nonumber \\
&+& \gamma'(I_{{\rm n}}(t_{d})+I_{{\rm s}}(t_{d})) \Delta t.
\label{vnn}
\end{eqnarray}
\begin{eqnarray}
s_{{\rm nnn}}(t_{d+1}|\mbox{\boldmath{$\theta$}},\gamma') &\approx& \frac{3}{2} \{ (\alpha+\beta)^{2} I_{{\rm n}}(t_{d}) + \gamma (\alpha+\beta)(2I_{{\rm n}}(t_{d})+I_{{\rm a}}(t_{d})) \} \Delta t^{2} \nonumber \\
&+& [ \frac{3}{2} \gamma' \{ (\alpha+\beta)(2I_{{\rm n}}(t_{d})+I_{{\rm s}}(t_{d})) + \gamma (2I_{{\rm n}}(t_{d}) + I_{{\rm a}}(t_{d}) + I_{{\rm s}}(t_{d})) \} + O(\gamma'^{2}) ] \Delta t^{2}. \nonumber \\
\label{snnn}
\end{eqnarray}
\begin{eqnarray}
\kappa_{{\rm nnnn}}(t_{d+1}|\mbox{\boldmath{$\theta$}},\gamma') &\approx& 3 \{(\alpha+\beta)^{3} I_{{\rm n}}(t_{d}) + \gamma (\alpha+\beta)^{2}(3I_{{\rm n}}(t_{d})+I_{{\rm a}}(t_{d})) + \gamma^{2}(\alpha+\beta)(I_{{\rm n}}(t_{d})+I_{{\rm a}}(t_{d})) \} \Delta t^{3} \nonumber \\
&+& [\gamma' \{ 3(\alpha+\beta)^{2}(3I_{{\rm n}}(t_{d})+I_{{\rm s}}(t_{d})) + 6\gamma(\alpha+\beta) (3I_{{\rm n}}(t_{d}) + I_{{\rm a}}(t_{d}) + I_{{\rm s}}(t_{d})) \nonumber \\
&+& \gamma^{2} (3I_{{\rm n}}(t_{d}) + I_{{\rm a}}(t_{d}) + 2I_{{\rm s}}(t_{d})) \} + O(\gamma'^{2}) ] \Delta t^{3}.
\label{knnnn}
\end{eqnarray}

Eq.(\ref{m(deltat)}) through (\ref{k(deltat)}) become eq.(\ref{ma}) through (\ref{kaaaa}) for $n_{{\rm a}}$.
\begin{eqnarray}
m_{{\rm a}}(t_{d+1}|\mbox{\boldmath{$\theta$}},\gamma') \approx I_{{\rm a}}(t_{d}) + \{(\alpha-\beta)I_{{\rm a}}(t_{d}) - \gamma(I_{{\rm a}}(t_{d})-I_{{\rm n}}(t_{d})) \} \Delta t.
\label{ma}
\end{eqnarray}
\begin{eqnarray}
v_{{\rm aa}}(t_{d+1}|\mbox{\boldmath{$\theta$}},\gamma') \approx \{(\alpha+\beta)I_{{\rm a}}(t_{d}) + \gamma(I_{{\rm a}}(t_{d})+I_{{\rm n}}(t_{d})) \} \Delta t.
\label{vaa}
\end{eqnarray}
\begin{eqnarray}
s_{{\rm aaa}}(t_{d+1}|\mbox{\boldmath{$\theta$}},\gamma') \approx \frac{3}{2} \{ (\alpha+\beta)^{2} I_{{\rm a}}(t_{d}) + \gamma (\alpha+\beta)(2I_{{\rm a}}(t_{d}) + I_{{\rm n}}(t_{d})) \} \Delta t^{2}.
\label{saaa}
\end{eqnarray}
\begin{eqnarray}
\kappa_{{\rm aaaa}}(t_{d+1}|\mbox{\boldmath{$\theta$}},\gamma') &\approx& 3 \{ (\alpha+\beta)^{3} I_{{\rm a}}(t_{d}) + \gamma (\alpha+\beta)^{2}(3I_{{\rm a}}(t_{d})+I_{{\rm n}}(t_{d})) + \gamma^{2}(\alpha+\beta)(I_{{\rm a}}(t_{d})+I_{{\rm n}}(t_{d})) \} \Delta t^{3} \nonumber \\
&+& \gamma' \gamma^{2} (I_{{\rm n}}(t_{d}) + I_{{\rm s}}(t_{d})) \Delta t^{3}.
\label{kaaaa}
\end{eqnarray}

\section{Measured moment}
\label{AppC}
The moments of $z_{i}$ are measured under the same experimental conditions as those for Figure \ref{022301s} {\bf a}, {\bf b}, {\bf e}, {\bf f}. The parameter $\mbox{\boldmath{$\theta$}}$ is either given, or unknown and estimated from the dataset. The following table shows the measured moments when a missing spreader node is absent. The values are the average over the all nodes and the trials for 100 different random network topologies.
\begin{center}
\begin{tabular}{|c|c|c|c|c|}
\hline
Parameter & $\langle m_{i} \rangle$ & $\langle v_{ii} \rangle$ & $\langle s_{iii} \rangle$ & $\langle \kappa_{iiii} \rangle$ \\
\hline
Given & 0.088 & 1.0 & 0.15 & 0.22 \\
\hline
Unknown & 0.043 & 1.0 & 0.11 & 0.062 \\
\hline
\end{tabular}
\end{center}

The following table shows the measured moments when the missing spreader node $n_{10}$ is an index node.
\begin{center}
\begin{tabular}{|c|c|c|c|c|c|}
\hline
\multicolumn{2}{|l|}{Parameter} & $\langle m_{i} \rangle$ & $\langle v_{ii} \rangle$ & $\langle s_{iii} \rangle$ & $\langle \kappa_{iiii} \rangle$ \\
\hline
Given & Neighboring node & 1.2 & 1.8 & 2.2 & 9.8 \\
\cline{2-6}
 & Non-neighboring node & -0.016 & 1.2 & 0.72 & 2.2 \\
\hline
Unknown & Neighboring node & 0.19 & 0.92 & 1.3 & 4.6 \\
\cline{2-6}
 & Non-neighboring node & 0.071 & 0.78 & 0.39 & 0.71 \\
\hline
\end{tabular}
\end{center}

The following table shows the measured moments when $n_{10}$ is an intermediate node.
\begin{center}
\begin{tabular}{|c|c|c|c|c|c|}
\hline
\multicolumn{2}{|l|}{Parameter} & $\langle m_{i} \rangle$ & $\langle v_{ii} \rangle$ & $\langle s_{iii} \rangle$ & $\langle \kappa_{iiii} \rangle$ \\
\hline
Given & Neighboring node & 0.61 & 1.2 & 0.062 & 0.13 \\
\cline{2-6}
 & Non-neighboring node & -0.16 & 1.0 & 0.17 & 0.30 \\
\hline
Unknown & Neighboring node & 0.11 & 1.1 & 0.056 & 0.011 \\
\cline{2-6}
 & Non-neighboring node & 0.0087 & 1.0 & 0.16 & 0.10 \\
\hline
\end{tabular}
\end{center}

\end{document}